\documentclass[letterpaper, 10 pt, conference]{ieeeconf}  

\IEEEoverridecommandlockouts                              

\usepackage[dvipsnames]{xcolor}

\usepackage{amsthm}
\usepackage{amsmath}
\usepackage[font=scriptsize,labelfont=bf]{caption}
\usepackage[font=scriptsize,labelfont=bf]{subcaption}
\usepackage{float}
\usepackage{algorithm}
\usepackage{algpseudocode}
\usepackage{multicol}
\usepackage{duckuments}
\usepackage{graphicx}
\usepackage{xr}
\let\labelindent\relax
\usepackage{enumitem}
\usepackage{etoolbox}
\usepackage{mathrsfs}
\usepackage{multirow}
\usepackage{flushend}

\DeclareCaptionType{CapEq}[][]
\definecolor{tab:blue}{RGB}{31,119,180}
\definecolor{tab:orange}{RGB}{255,127,14}
\definecolor{tab:green}{RGB}{44,160,44}
\definecolor{tab:red}{RGB}{214,39,40}
\definecolor{tab:purple}{RGB}{148,103,189}

\definecolor{sns:blue}{rgb}{0.00392156862745098, 0.45098039215686275, 0.6980392156862745}
\definecolor{sns:orange}{rgb}{0.8705882352941177, 0.5607843137254902, 0.0196078431372549}
\definecolor{sns:green}{rgb}{0.00784313725490196, 0.6196078431372549, 0.45098039215686275}
\definecolor{sns:red}{rgb}{0.8352941176470589, 0.3686274509803922, 0.0}
\definecolor{sns:pink}{rgb}{0.8, 0.47058823529411764, 0.7372549019607844}
\definecolor{sns:brown}{rgb}{0.792156862745098, 0.5686274509803921, 0.3803921568627451}
\definecolor{sns:gray}{rgb}{0.5803921568627451, 0.5803921568627451, 0.5803921568627451}
\newtheoremstyle{dense}
  {3pt} 
  {3pt} 
  {\itshape} 
  {} 
  {\bfseries} 
  {:} 
  {.5em} 
  {} 
\theoremstyle{dense}

\newtheorem{remark}{Remark}
\AtBeginEnvironment{remark}{\small}
\AtBeginEnvironment{quote}{\par\singlespacing\small}
\AtBeginEnvironment{tabular}{\footnotesize}
\IEEEaftertitletext{\vspace{-1\baselineskip}}
\usepackage{stix}
\usepackage{tikz}
\DeclareRobustCommand{\thickX}{
    \begin{tikzpicture}[baseline=0ex, line width=2, scale=0.13];
    \draw (0,0) -- (1,1);
    \draw (0,1) -- (1,0);
    \end{tikzpicture}}
\newcommand{\SymPcm}{\textcolor{sns:green}{$\thickX$}}
\newcommand{\SymRisam}{\textcolor{sns:blue}{$\bigstar$}}
\newcommand{\SymGnc}{\textcolor{sns:orange}{$\mdblksquare$}}
\newcommand{\SymGm}{\textcolor{sns:red}{$\blacktriangledown$}}
\newcommand{\SymMaxMix}{\textcolor{sns:pink}{$\mdblkcircle$}}
\newcommand{\SymHuber}{\textcolor{sns:brown}{$\blacklozenge$}}
\newcommand{\SymSC}{\textcolor{sns:gray}{$\pentagonblack$}}

\makeatletter
\let\NAT@parse\undefined
\makeatother
\usepackage{hyperref}
\usepackage{cleveref}
\hypersetup{
    colorlinks=true,
    linkcolor=blue,
    filecolor=magenta,      
    urlcolor=cyan,
    pdfpagemode=FullScreen,
}

\title{\LARGE \bf Robust Incremental Smoothing and Mapping (riSAM)\vspace{-2ex}
\author{Daniel McGann$^1$, John G. Rogers III$^2$, and Michael Kaess$^1$}
\thanks{\scriptsize
This work was partially supported by DEVCOM Army Research Laboratory Distributed, Collaborative, Intelligent Systems and Technology Collaborative Research Alliance (DCIST-CRA) W911NF-17-2-0181 and by NSF grant IIS-2008279.}
\thanks{ \scriptsize
$^1$ D. McGann and M. Kaess are with the Robotics Institute, Carnegie Mellon University, Pittsburgh, PA, USA. \texttt{\{danmcgann, kaess\}@cmu.edu}
}
\thanks{\scriptsize
$^2$ J. Rogers is with the DEVCOM Army Research Laboratory, Adelphi, MD, USA. \texttt{john.g.rogers59.civ@army.mil}
}
}

\begin{document}

\maketitle
\thispagestyle{empty}
\pagestyle{empty}

\begin{abstract}
This paper presents a method for robust optimization for online incremental Simultaneous Localization and Mapping (SLAM). Due to the NP-Hardness of data association in the presence of perceptual aliasing, tractable (approximate) approaches to data association will produce erroneous measurements. We require SLAM back-ends that can converge to accurate solutions in the presence of outlier measurements while meeting online efficiency constraints. Existing robust SLAM methods either remain sensitive to outliers, become increasingly sensitive to initialization, or fail to provide online efficiency. We present the robust incremental Smoothing and Mapping (riSAM) algorithm, a robust back-end optimizer for incremental SLAM based on Graduated Non-Convexity. We demonstrate on benchmarking datasets that our algorithm achieves online efficiency, outperforms existing online approaches, and matches or improves the performance of existing offline methods.


\end{abstract}

\section{Introduction}\label{sec:intro}
\vspace{-2pt}

Simultaneous Localization and Mapping (SLAM) is a key component that supports many complex robot behaviors. The modern approach to SLAM is to separate the task into a front-end process responsible for parsing raw sensor data into useful measurements and a back-end process responsible for estimating unknown variables from these measurements. A large variety of front-end processes have been proposed for various sensing modalities~\cite[Sec.~B]{leonard_slam_survey_2016}. However, all front-ends must solve the data association problem. This is of particular importance given that the complexity of data association, in the presence of perceptual aliasing, is NP-Hard~\cite{lajoie_dc_convex_opt_2019}. Therefore, in order to maintain tractability, we can only solve this association approximately. 

Unlike front-end processes, the SLAM community has converged to a single standard approach to the back-end. This approach formulates SLAM as Maximum a Posteriori (MAP) estimation and solves for the estimate using local Nonlinear Least Squares (NLS) optimization~\cite{leonard_slam_survey_2016, factor_graphs_for_robot_perception}. Importantly, back-end methods have been developed that can operate both \textit{incrementally} and \textit{online}, which is necessary for application to real-world robotics~\cite[Sec.~2]{isam2}

Since NLS optimization is sensitive to outliers, imperfect front-end processes can cause poor performance in SLAM systems that are unable to handle these outliers. Given the importance of SLAM in robotics, techniques to handle outliers have been well studied (See Sec.~\ref{sec:related-work}). These methods have shown promising results, however, state-of-the-art robust SLAM back-ends are predominantly batch methods. While batch methods can be used in place of incremental ones, their complexity permits online operation only for small problems. Additionally, existing incremental algorithms that can achieve online efficiency on moderate and large problems remain brittle under common conditions making them difficult to apply generally (See Sec.~\ref{sec:experiments}). 

In this paper, we present a novel incremental algorithm for robust SLAM that achieves performance equal to or better than that of state-of-the-art batch methods while maintaining online efficiency. As part of this algorithm we develop (1) an efficient form of Graduated Non-Convexity~\cite{yang_gnc_2020}, (2) the Scale Invariant Graduated kernel which provides improved performance and efficiency, and (3) a novel trust region optimization algorithm for incremental graduated optimization. Finally, we evaluate our algorithm under a variety of conditions on benchmark datasets demonstrating its ability to outperform existing methods.

\begin{figure}[t!]
    \vspace{-2pt}
    \centering
    \begin{subfigure}{0.45\linewidth}
      \centering
      \includegraphics[width=\linewidth]{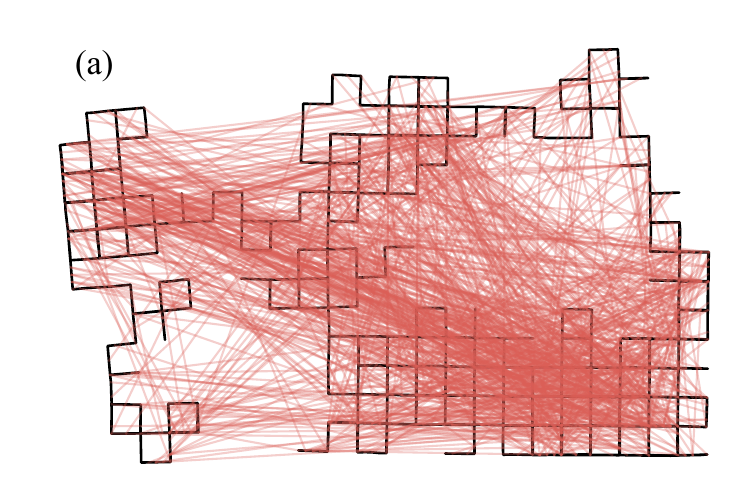}
      \label{fig:example_trajectory_manhattanWorld}
      \vspace{-0.6cm}
    \end{subfigure}
    \begin{subfigure}{0.45\linewidth}
      \centering
      \includegraphics[width=\linewidth]{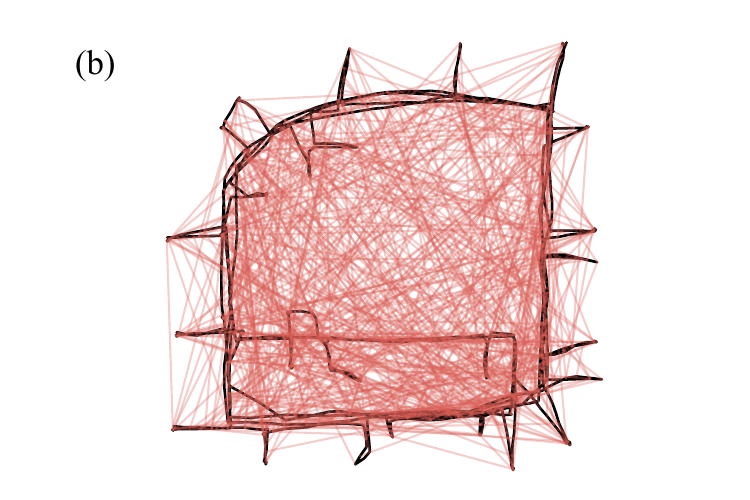}
      \label{fig:example_trajectory_sphere}
      \vspace{-0.6cm}
    \end{subfigure}
    \begin{subfigure}{0.45\linewidth}
      \centering
      \includegraphics[width=\linewidth]{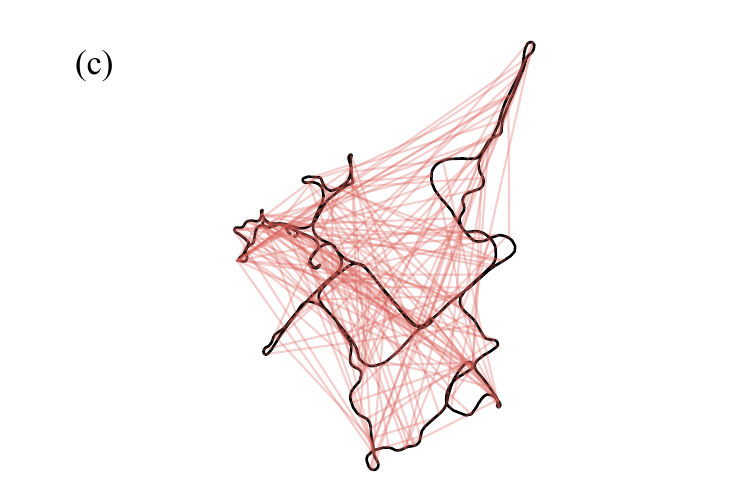}
      \label{fig:example_trajectory_city10k}
    \end{subfigure}
    \begin{subfigure}{0.45\linewidth}
      \centering
      \includegraphics[width=\linewidth]{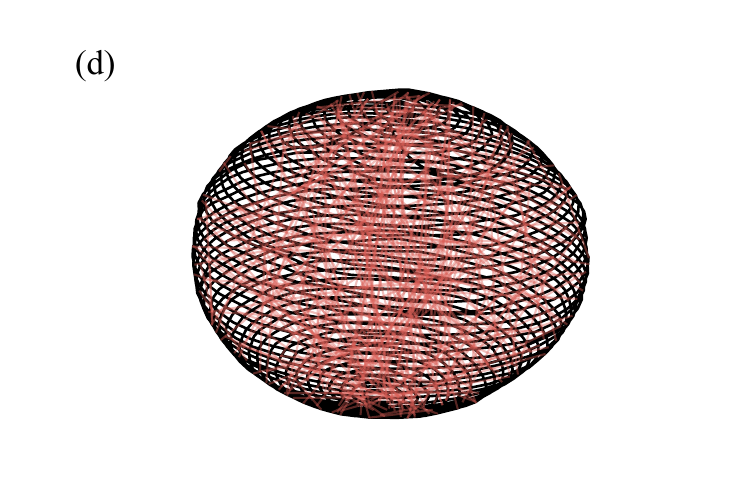}
      \label{fig:example_trajectory_garage}
    \end{subfigure}
    \vspace{-0.5cm}
    \caption{Examples of riSAM solutions on the \textbf{(a)} Manhattan 3500 \textbf{(b)} Intel  \textbf{(c)} CSAIL and \textbf{(d)} Sphere 2500 datasets corrupted with outlier loop closures (shown in red).
    }
    \label{fig:example_risam_solutions}
\end{figure}

\section{Related Work}\label{sec:related-work}
\vspace{-2pt}
Approaches to robust SLAM can be broadly classified as either maximum consensus or robust estimation methods. 

Maximum consensus methods attempt to sanitize incoming measurements by identifying the largest set that are jointly consistent. They then apply existing non-robust optimization techniques assuming only inlier measurements remain. While it is possible to exactly compute the the maximally consistent set, doing so requires exponential time. Exact methods like Joint Compatibility Branch and Bound (JCBB) show that algorithms can often do better than exponential time, but are still too inefficient for real-time constraints~\cite{neira_jcbb_2001}. For tractability, other methods seek to approximate the maximum consensus set. Pairwise Consistency Maximization (PCM) does so by approximating joint consistency with pairwise consistency and using approximate Max-Clique algorithms to compute the maximum pairwise consistent set~\cite{mangelson_pcm_2018}. Realizing, Reversing, and Recovering (RRR) and its incremental derivative (iRRR) approximate the maximum consensus set using repeated optimization of temporally partitioned sets of measurements~\cite{latif_rrr_irrr_2013, latif_irrr_2012}. However, despite these approximations, neither method claims real-time efficiency and as we will see in Sec.~\ref{sec:experiments} these approximations may not always result in good performance.

Robust estimation methods attempt to construct the SLAM problem such that the effects of outlier measurements on the final solution are mitigated. The oldest and most common form of robust estimation is the use of M-Estimators~\cite{zhang_conic_1997}. This method modifies the NLS cost function with a robust kernel $\rho$ that applies a sub-quadratic cost to outliers. This cost function can be seen in Eq.~\eqref{eq:robust_kernel_cost_landscape}:
\begin{equation}
    \label{eq:robust_kernel_cost_landscape}
    \sum \rho\left({r\left(z_i, f\left(x_i\right)\right)}\right)
\end{equation}
where the function $r$ computes the residual between the $i^{th}$ measurement ($z_i$) and its predicted value $(f(x_i)$) given a variable estimate ($x_i$). This robust cost function is solvable with Iteratively Reweighted Least Squares (IRLS) when the robust kernel meets known criteria~\cite{aftab_mest_convergence_2015}. A large number of different robust kernels have been proposed~\cite{zhang_conic_1997, agarwal_dcs_2013, olson_max_mix}. Unfortunately, M-Estimator optimization is very sensitive to the choice of $\rho$. Kernels with infinite growth (e.g. Huber) allow outliers to have significant influence on the solution. Conversely, kernels that are asymptotically constant (e.g. Geman-McClure) introduce local optima to the cost function and cause sensitivity to initialization~\cite[A6.8]{hartley_multiple_2003}. Methods have been proposed to better optimize the cost function when using asymptotically constant kernels. Graduated Non-Convexity (GNC) applies ideas from continuation to better optimize this highly non-convex cost function~\cite{yang_gnc_2020}. Empirical evidence shows that using GNC significantly reduces sensitivity to initialization that such kernels typically exhibit.

Another robust estimation approach is to augment the SLAM problem with additional variables that, during optimization, classify measurements and reduce the influence of outliers. Switchable Constraints adds continuous variables for each measurement that softly classify measurements as inliers or outliers~\cite{sunderhauf_switchable_2012}. Interestingly, this approach has been shown to be equivalent to the use of M-Estimators when the switch variables are conditionally independent~\cite{black_Rangarajan, yang_gnc_2020}. Thus, we expect it to exhibit the same behavior as M-Estimators. In a similar manner, one can augment the graph with discrete variables to classify measurements and solve the mixed problem using alternating optimization~\cite{doherty_dcsam_2022}. Similar to the continuous case, when these variables are conditionally independent, this is equivalent to the use of a robust kernel (a Max-Mixture Kernel~\cite{olson_max_mix} in the case of the approach presented by Doherty et al.~\cite{doherty_dcsam_2022}). 

Variable augmentation can also allow for more complex modeling of correlations between measurements. However, solving the resulting graphical models can be difficult due to their hybrid nature or decreased sparsity. Prior works have used convex relaxations to find solutions and even provide correctness guarantees~\cite{lajoie_dc_convex_opt_2019}. Unfortunately, these methods require significant computational time and cannot meet online constraints. Another alternative model is employed by the Adaptive Kernel method which adds a single continuous variable to the optimization~\cite{chebrolu_adaptive_2021}. This variable optimizes the shape of the robust kernel used for all measurements to avoid the sensitivity derived from hand selecting a kernel. However, after the kernel shape converges the method recovers that of a standard M-Estimator resulting, once again, in sensitivity to either outliers or initialization. 

Robust estimation methods are generally compatible with any NLS solver including existing fast and efficient incremental SLAM solvers like iSAM2~\cite{isam2, dellaert_gtsam_tech_report_2012}. Therefore, unlike maximum consensus methods, some robust estimation methods can achieve online efficiency for moderate to large problems. The exceptions to this rule are approaches that solve via convex relaxations or GNC. These methods run offline and in batch~\cite{lajoie_dc_convex_opt_2019, yang_gnc_2020}.

Given currently published results and further supported in Sec~\ref{sec:experiments}, it appears that out of all robust SLAM methods GNC can provide the best results across general SLAM scenarios. We pursue this approach to robust SLAM and introduce an incremental version of GNC that is able to match the performance of its batch variant, outperform alternative methods, and achieve online efficiency. 
\section{Methodology}\label{sec:methodolody}
\vspace{-2pt}
In this section we present the \textit{robust, incremental Smoothing and Mapping} (riSAM) algorithm. We first present the individual building blocks of riSAM before summarizing the algorithm in full. 

\subsection{Graduated Non-Convexity (GNC)} \label{sec:prelim:gnc}
\vspace{-2pt}
GNC is a technique that seeks to mitigate the difficulties observed when optimizing the robust cost function (Eq.~\eqref{eq:robust_kernel_cost_landscape})~\cite{yang_gnc_2020}. It does so by first convexifying the problem and then solving a series of progressively less convex instances. GNC uses the solutions from more convex problems to initialize the less convex versions. The degree of convexity is changed by a control parameter $\mu$ which affects the shape of a graduated robust kernel $\rho(r;\mu)$. We refer to this general process as "graduation" though it is also known in optimization literature as "continuation" and is frequently employed to optimize difficult problems~\cite{carlone_estimation_2022}.  

\begin{figure}[H]
    \centering
    \includegraphics[width=\linewidth]{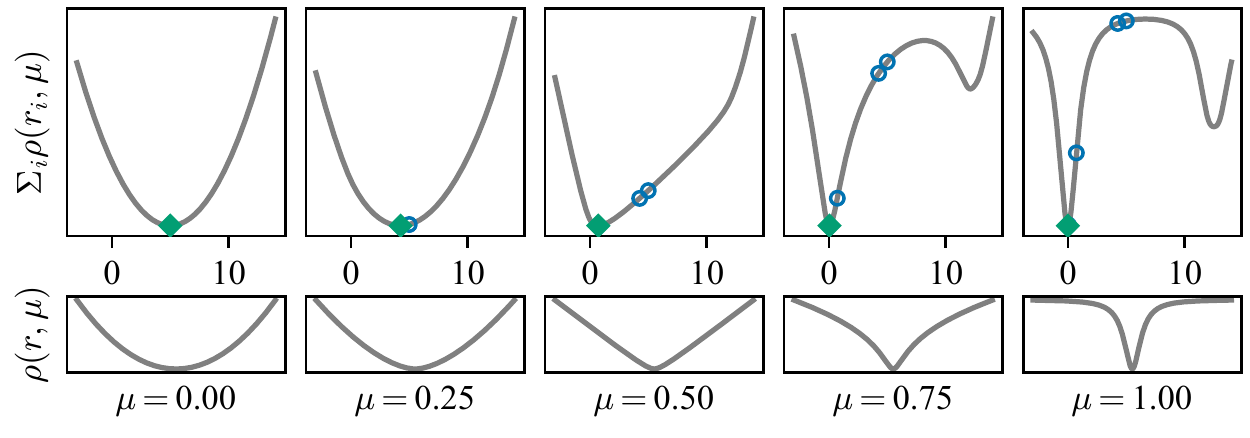}
    \vspace{-0.7cm}
    \caption{Example of GNC solving a linear problem with outliers $\sum_i \rho(||z_i||_2^2, \mu)~\forall~ z_i\in\{0.1, -0.05, -0.1, 12, 13\}$. The top row depicts the cost function while the bottom row depicts the shape of the robust kernel for each $\mu$. Diamonds (\textcolor{sns:green}{$\mdlgblkdiamond$}) mark the global optima for each cost function and circles (\textcolor{sns:blue}{$\bigcirc$}) indicate the initialization history. Initialization from solving the convexified problem (left) ensure that we converge to the global optima for the non-convex problem (right). 
    }
    \label{fig:gnc_demonstration}
\end{figure}

We can understand how graduation helps solve the robust SLAM problem by relating it to the behaviors of different robust kernels discussed in Sec.~\ref{sec:related-work}. For a \textit{linear problem}, we know that infinite-growth kernels will result in cost functions with a single well defined optimum that is skewed by outliers. Inversely, asymptotically constant kernels result in a cost function with many local optima, but where the global optimum well approximates the outlier-free case. Using GNC we better initialize the non-convex versions of the problem reducing the likelihood that we converge to a local-optima. While this is evident for a linear problem (Fig~\ref{fig:gnc_demonstration}), we find the same behavior even in nonlinear problems like SLAM.

\subsection{Efficient GNC} \label{sec:methodology:efficien_gnc}
\vspace{-2pt}

GNC provides an effective method to optimize the robust cost function at a computational cost. GNC, as presented by Yang et al. \cite{yang_gnc_2020}, requires we solve the global optima of the cost function for each value of $\mu$. For SLAM, this requires an iterative optimization routine, making the process inefficient. However, for all values of $\mu$ except the final, we need iterate only until the variable estimates lie within the basin of convergence of the cost function defined by the next $\mu$ value. While, it is impractical to evaluate when this condition is met, full convergence is likely unnecessary. For efficiency, we propose using only a single update per control parameter value (Alg.~\ref{alg:efficient_gnc}). Experimental results (Sec.~\ref{sec:experiments}) indicate that for typical SLAM problems this approximation is sufficient.

\begin{algorithm}
\footnotesize 
\captionsetup{font=footnotesize} 
\caption{Efficient Graduated Non-Convexity}\label{alg:efficient_gnc}
    \begin{algorithmic}[1]
    \State In: Nonlinear Problem $\mathcal{P}$, Initial Variable Estimate $\mathcal{X}_0$, Control Parameter Bounds $\mu_{init}$ and $\mu_{final}$
    \State Out: Final Variable Estimate $\mathcal{X}_{final}$
    \State $\mu_0 \gets \mu_{init}$
    \While{Not IsConverged($\mu_i, \mu_{final}$)}
        \State $\mathcal{L} \gets \text{Linearize}(\mathcal{P}, \mathcal{X}_i, \mu_i)$ \label{alg:efficient_gnc:linearize}
        \State $\mathcal{R}, b \gets \text{Solve}(\mathcal{L})$ \label{alg:efficient_gnc:solve}
        \State $\Delta \gets \text{UpdateStepAlgorithm}(\mathcal{R}, b)$
        \State $\mathcal{X}_{i+1} \gets \mathcal{X}_i \oplus \Delta$
        \State $\mu_{i+1} \gets$ Update($\mu_i$)
    \EndWhile
    \end{algorithmic}
\end{algorithm}

\subsection{The Scale Invariant Graduated (SIG) Kernel} \label{sec:methodology:sigk}
\vspace{-2pt}
The authors of GNC propose two graduated kernels based on the Geman-McClure (GM)~\cite[Eq.~4]{yang_gnc_2020} and the Truncated Least Squares (TLS)~\cite[Eq.~6]{yang_gnc_2020} kernels. Both kernels transition between a quadratic kernel and their respective basis kernel for $\mu \in [0, \infty)$. Thus, for these kernels, it takes theoretically infinite steps to transition between the convex and non-convex cases. In practice, a range can be constructed to admit a finite number of steps for $\mu$ to converge from $\mu_{init}$ to $\mu_{final}$~\cite[Remark.~5]{yang_gnc_2020}. However, the number of steps is dependent on the largest measurement residual. This would result in an inconsistent number of iterations within Alg.~\ref{alg:efficient_gnc}. Also, such a construction admits theoretically incorrect behavior as the convexity structure of GNC-GM and GNC-TLS will not be consistent across the residual domain for any value of $\mu$. Therefore, using these kernels allows variable estimates to jump into non-convex regions of the cost function even for iterations of GNC in which we expect the problem to be fully convex.

\begin{figure}[h]
    \centering
    \begin{subfigure}{0.24\textwidth}
      \centering
      \includegraphics[width=\linewidth]{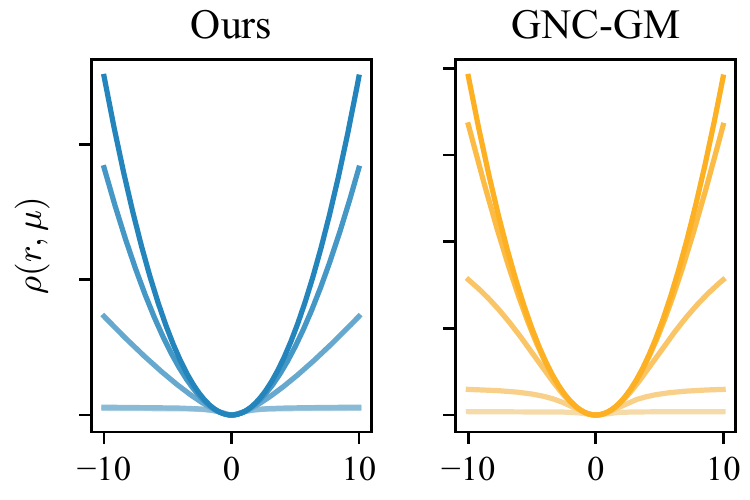}
      \caption{Small Scale}
      \label{fig:sigk:small_scale}
    \end{subfigure}%
    \begin{subfigure}{0.24\textwidth}
      \centering
      \includegraphics[width=\linewidth]{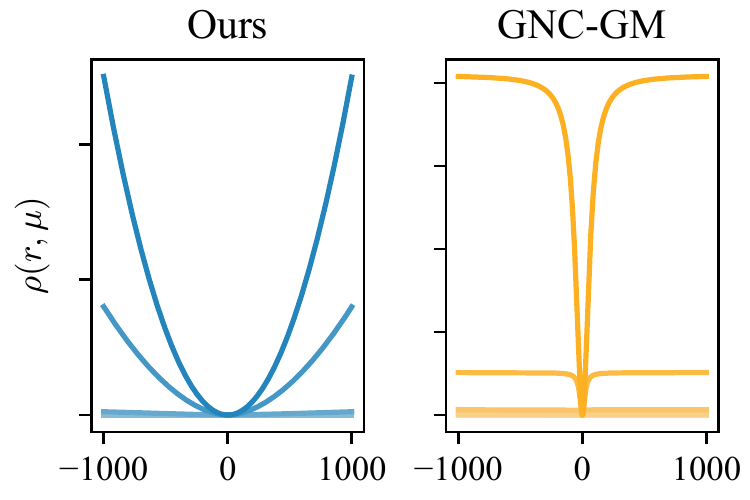}
      \caption{Large Scale}
      \label{fig:sigk:large_scale}
    \end{subfigure}%
\caption{The SIG kernel compared to the GNC-GM kernel~\cite{yang_gnc_2020}  at two different scales. While the GNC-GM kernel appears to transition from convex to non-convex at the smaller scale (\ref{fig:sigk:small_scale}) the kernel is not actually convex across the entire domain (\ref{fig:sigk:large_scale}). Our kernel is convex over a scale invariant domain for $\mu \geq 0.5$ and non-convex for $\mu < 0.5$.}
    \label{fig:scale_invariant_kernel_comparison}
\end{figure}

For online efficiency, we require a kernel that admits a known constant number of GNC iterations. Further, we desire a kernel for which convexity is known to be invariant to the scale of the residual domain. Towards this goal, we propose the Scale Invariant Graduated (SIG) kernel (Eq.~\eqref{eq:scale_invariant_kernel}).
\begin{equation}
    \rho_{\text{SIG}}(r;\mu) = \frac{1}{2} \frac{c^2 r^2}{c^2 + (r^2)^\mu}
    \label{eq:scale_invariant_kernel}
\end{equation}
Based on the Geman-McClure kernel, the SIG kernel is quadratic when $\mu=0$ and recovers the Geman-McClure kernel when $\mu=1$. The shape parameter $c$ defines the width of quadratic support of the basis kernel. The finite range of $\mu$ enables us to perform a known constant number of iterations in Alg.\ref{alg:efficient_gnc}. Additionally, the convexity structure of this kernel is invariant to the scale of the domain and is known for ranges of $\mu$ as shown in Fig.~\ref{fig:scale_invariant_kernel_comparison}.

\begin{remark}[SIG Kernel Impl. Details]\label{remark:sigk_impl_details}
We update $\mu$ according to $\mu_{i+1} = \min\left(1.0, \mu_i + 1.2(\mu_i - \mu_{init} + 0.1) \right)$. This update rule was found empirically using the intuition that we desire multiple convex steps and multiple non-convex steps before convergence. The shape of the kernel for these values can be seen in Fig.~\ref{fig:scale_invariant_kernel_comparison}. We select $c=3$ as we expect inliers fall within $3\sigma$ of the mean.
\end{remark}

\subsection{Graduated iSAM2} \label{sec:methodology:graduated_isam2}
\vspace{-2pt}
While Alg.~\ref{alg:efficient_gnc} and Eq.~\eqref{eq:scale_invariant_kernel} reduce the computational cost of GNC, it is still a batch process. We, however, can incrementalize Alg.~\ref{alg:efficient_gnc} as the algorithm is agnostic to the process used in Steps \ref{alg:efficient_gnc:linearize} and \ref{alg:efficient_gnc:solve} to linearize and solve the problem. We can therefore use existing incremental SLAM solvers in conjunction with our Efficient GNC algorithm.

Specifically, we adapt the iSAM2 algorithm for this purpose~\cite{isam2}. We modify iSAM2's update procedure~\cite[Alg. 6]{isam2} to linearize according to the SIG kernel and GNC control parameter $\mu$. For correctness, we also modify the algorithm to track all variables affected by factors that are relinearized with $\mu \neq \mu_{final}$ (See Remark~\ref{remark:risam_correctness}). The modified version can be found in Alg.~\ref{alg:graduated_isam2_update} with modified steps marked in bold. 

\begin{algorithm}
\footnotesize 
\captionsetup{font=footnotesize} 
\caption{Graduated iSAM2 Update}\label{alg:graduated_isam2_update}
    \begin{algorithmic}[1]
    \State In: Bayes Tree $\mathscr{T}$ , Nonlinear Factors $\mathscr{F}$, Affected Variables $\mathscr{J}$, Control Parameter $\mu$
    \State Out: Modified Bayes Tree $\mathscr{T}'$, Convex Variables $\mathscr{C}$
    \State Remove top of Bayes tree:
    \begin{enumerate}[label=(\alph*)]
        \item For each affected variable in $\mathscr{J}$ remove the corresponding
    clique and all parents up to the root.
        \item{Store orphaned sub-trees $\mathscr{T}_{orph}$ of removed cliques.}
    \end{enumerate}
    \State \textbf{Relinearize all factors required to recreate top, using $\mu$ if applicable}.
    \State \textbf{Store variables of factors relinearized with $\mu \neq \mu_{final}$ in $\mathscr{C}$}.
    \State Add cached linear factors from orphans $\mathscr{T}_{orph}$.
    \State Re-order variables, see Section 3.4.
    \State Eliminate factor graph~\cite[Alg. 2]{isam2}, create new Bayes tree~\cite[Alg. 3]{isam2}.
    \State Insert the orphans $\mathscr{T}_{orph}$ back into the new Bayes Tree.
    \end{algorithmic}
\end{algorithm}

\subsection{Dog-Leg Line Search} \label{sec:methodology:dogleg_line_search}
\vspace{-2pt}
Within Alg.~\ref{alg:efficient_gnc} we compute updates to the NLS problem. Existing SLAM solvers often use the Gauss-Newton algorithm to compute updates. However, we find that using Gauss-Newton on datasets corrupted with outliers causes large updates and results in numerical overflow. The same behavior is observed in Alg.~\ref{alg:efficient_gnc} during interations where $\rho(r; \mu)$ is convex.

To handle such behavior we turn to trust region optimization algorithms. The only trust region algorithm that has been found to be incrementalizable is Powell's Dog-Leg, whose incremental variant is the RISE algorithm~\cite{rosen_rise_2014}. RISE, however, is incompatible with our graduated approach. Firstly, changes in $\mu$ mean that the trust region is not correlated between iterations as expected by RISE. Secondly, RISE's step-acceptance criterion relies on comparing the cost decrease between the nonlinear and linearized problems. We find this criteria is biased to accept steps where $\rho(r; \mu)$ is convex and reject steps where $\rho(r; \mu)$ is non-convex.

Given that no viable method exists, we developed a novel trust region algorithm tailored to GNC that, like RISE, is based on Powell's Dog-Leg. It is structured as a standard line-search algorithm satisfying the Wolfe conditions~\cite{nocedal_numopt_2006} with three modifications. Firstly, we set a maximum step size $\alpha_{max}$ to avoid numerically unstable steps. Secondly, we search along the Dog-Leg arc rather than along a single direction to gain the advantages of combining the Gauss-Newton and gradient directions. Thirdly, we always accept the first step regardless of whether it satisfies convergence conditions to encourage the exploration of the cost function.

\begin{algorithm}
\footnotesize 
\captionsetup{font=footnotesize} 
\caption{Dog-Leg Line Search}\label{alg:dogleg_line_search}
    \begin{algorithmic}[1]
    \State In: Bayes Tree $\mathscr{T}$, Gauss-Newton Step $\Delta_{GN}$, Gradient Step $\Delta_{G}$
    \State Out: Update Step $\Delta$
    \State $\alpha_0 = \min{\left(\alpha_{min}, |\Delta_{GN}|\right)}$ \Comment{Initial trust region size.}
    \State $\alpha_f = \min{\left(\alpha_{max}, |\Delta_{GN}|\right)}$
    \Comment{Maximum trust region size.}
    \State $\Delta \gets \text{ComputeDogLegPoint}(\mathscr{T}, \Delta_{GN}, \Delta_{G}, \alpha_0)$\Comment{\cite[Alg.~3]{rosen_rise_2014}}
    
    \While{$\alpha_{i+1}=\text{Update}(\alpha_i)\leq\alpha_f$ and Wolfe cond. are not satisfied}
    \State $\Delta_{test} \gets \text{ComputeDogLegPoint}(\mathscr{T}, \Delta_{GN}, \Delta_{G}, \alpha_i)$
    \State $\Delta \gets \Delta_{test}$ if Wolfe cond. are satisfied.
    \EndWhile
    \end{algorithmic}
\end{algorithm}

\begin{remark}[Dog-Leg Line Search Impl. Details]\label{remark:dlls_impl_details}
To allow for convergence, the initial step size and max step size are upper-bounded by the magnitude of the Gauss-Newton step. $\alpha_{min}$ is selected $ \approx 1$. $\alpha_{max}$ is selected based on the scale of the robot's operational environment (i.e. $O(10^2)$ for building sized environment). $\text{Update}(\alpha)$ is selected as $\alpha_{i+1} = 1.5 \alpha_i$. Finally, the user must supply a sufficient decrease coefficient for checking the Wolfe conditions.
\end{remark}

\subsection{riSAM} \label{sec:methodology:risam}
\vspace{-2pt}
Putting together Efficient GNC (Alg.~\ref{alg:efficient_gnc}), the SIG kernel (Eq.~\eqref{eq:scale_invariant_kernel}), our graduated iSAM2 update (Alg.~\ref{alg:graduated_isam2_update}), and the Dog-Leg Line Search Algorithm (Alg.~\ref{alg:dogleg_line_search}) we create the robust incremental Smoothing and Mapping (riSAM) algorithm. We summarize the algorithm in Alg.~\ref{alg:risam}

\begin{algorithm}
\footnotesize 
\captionsetup{font=footnotesize} 
\caption{riSAM}\label{alg:risam}
    \begin{algorithmic}[1]
    \State In: Bayes Tree $\mathscr{T}$, Variable Estimate $\mathcal{X}_0$, Nonlinear Factors $\mathscr{F}$, Affected Variables $\mathscr{J}$
    \State Out: Modified Bayes Tree $\mathscr{T}'$, Updated Variable Estimate $\mathcal{X}'$
    \State $\mu_0 \gets \mu_{init}$\Comment{$\mu_{init}=0$ Sec.~\ref{sec:methodology:sigk}}
    \While{not IsConverged$(\mu_i, \mu_{final})$}\Comment{$\mu_{final}=1$ Sec.~\ref{sec:methodology:sigk}}
        \State $\mathscr{T}, \mathscr{C} \gets$ UpdateBayesTree$(\mathscr{T}, \mathscr{F}, \mathscr{J}, \mu_i)$ \Comment{Alg.~\ref{alg:graduated_isam2_update}, Eq.~\eqref{eq:scale_invariant_kernel}}
        \State $\Delta \gets \text{DogLegLineSearch}(\mathscr{T}, \Delta_{GN}, \Delta_{G})$ \Comment{Alg.~\ref{alg:dogleg_line_search}}
        \State $\mathcal{X}_{i+1} \gets \mathcal{X}_i \oplus \Delta$
        \State $\mathscr{J} \gets \mathscr{C} ~\cup$ FluidRelinearization($\Delta$) \Comment{\cite[Alg.~5]{isam2}} 
    
        \State $\mu_{i+1} \gets$ UpdateMu$(\mu_i)$ \Comment{Remark~\ref{remark:sigk_impl_details}}
    \EndWhile
    \end{algorithmic}
\end{algorithm}

\begin{remark}[riSAM Correctness]\label{remark:risam_correctness}
New information is only added in the first Bayes tree update of the riSAM algorithm. To ensure that all relevant factors are included in subsequent updates to the Bayes tree and are therefore relinearized according to the updated value of $\mu$ we must track variables of these factors in Alg.~\ref{alg:graduated_isam2_update} and mark them in subsequent iterations. Additionally, to ensure correctness we account for fluid relinearization during internal iterations of the riSAM algorithm (For fluid relinearization details see~\cite[Sec.~4.1]{isam2}).
\end{remark}

While the base riSAM algorithm is a full incremental robust SLAM solver, we further improve its efficiency with some minor modifications. 

\subsubsection{Known Inliers} In-so-far we have treated all measurements identically. However, in practical applications there are likely measurements that we don't need consider as potential outliers (e.g. odometry). Within riSAM, these measurements can be given a quadratic kernel and when such a new measurement arrives we apply only a base iSAM2 update skipping the internal loop of Alg.~\ref{alg:risam} for efficiency.

\subsubsection{Adjusting Initial Control Parameters} Large numbers of outlier measurements have the effect of reducing the sparsity of the underlying SLAM problem (e.g. Fig.~\ref{fig:example_risam_solutions}). Given we rely on sparsity for our efficiency, this can greatly increase computational cost. However, we expect that over time we accumulate evidence that some measurements are definite outliers. We use this knowledge to mitigate the effect of outliers by maintaining a unique $\mu^i_{init}$ for each measurement $z_i$. When the algorithm converges with respect to its variable estimate we classify \textit{strong inliers} and \textit{strong outliers} according to user defined $\chi^2$ thresholds (we select thresholds of $0.25$ and $0.9$ respectively). For all strong outliers we adjust $\mu^i_{init}$ according to Remark~\ref{remark:sigk_impl_details}, and for all strong inliers we adjust their $\mu^i_{init}$ by $\max\left(0, \mu^i_{init} - 0.1\right)$. This results in fewer variables marked convex by Alg.~\ref{alg:graduated_isam2_update} and later iterations of Alg.~\ref{alg:risam} are, in turn, more efficient.

\section{Experiments}\label{sec:experiments}
\vspace{-2pt}
In this section we provide empirical evidence to support the effectiveness and efficiency of the riSAM algorithm when handling erroneous loop-closures that we expect due to perceptual aliasing. We evaluate against state-of-the-art batch SLAM algorithms GNC~\cite{yang_gnc_2020} and PCM~\cite{mangelson_pcm_2018, rosinol_kimera_2020} as well as against incremental robust estimation methods Switchable Constraints (SC)~\cite{sunderhauf_switchable_2012}, Max-Mixture kernels~\cite{olson_max_mix}, Gemen-McClure kernels~\cite{zhang_conic_1997}, and Huber kernels~\cite{zhang_conic_1997} which use iSAM2 as their underlying solver~\cite{isam2}. We also perform two ablation studies to evaluate the efficacy of our Dog-Leg Line Search algorithm and the efficiency gain from $\mu_{init}$ updates.

\begin{remark}[Equivalencies]
In Sec.~\ref{sec:related-work} we note that there are equivalencies between M-Estimators and corresponding variable augmentation methods. We investigated this equivalency for continuous variables and discrete variables. We found that in the case of discrete variables the variable augmentation method~\cite{doherty_dcsam_2022} was empirically equivalent to the kernel approach~\cite{olson_max_mix}. For clarity we report only the kernel method (Max-Mixtures). Interestingly, we found that in the continuous case this equivalency does not result in identical empirical behavior. For further discussion see Sec.~\ref{sec:discussion}.
\end{remark}

\begin{remark}[Prior Work Impl. Details]
GNC provided by GTSAM\footnote{\scriptsize\url{https://github.com/borglab/gtsam}} and uses default parameters. PCM provided by Kimera\footnote{\scriptsize\url{https://github.com/MIT-SPARK/Kimera-RPGO}}. Switchable Constraints provided by OpenSLAM\footnote{\scriptsize\url{https://github.com/OpenSLAM-org/openslam_vertigo}}  Max-Mixtures were implemented by the authors and use a isotropic outlier noise model with $\sigma = 10^7$ and an outlier weight of $10^{-7}$ based on the values used in the original work~\cite{olson_max_mix}. To match the riSAM shape parameter (See Remark~\ref{remark:sigk_impl_details}), we use $c=3$ for all kernel methods, a threshold of $3$ for PCM, and a switch variable covariance of $3$ for SC.
\end{remark}

\subsection{Metrics}
\vspace{-2pt}

We seek to evaluate the effectiveness of the algorithms at correctly rejecting outlier measurements and providing accurate solutions at every time-step. To quantify performance we propose the following incremental versions of Absolute Trajectory Error (ATE)~\cite{zhang_traj_metric_tutorial_2018}, Precision, and Recall:
\begin{equation}
\label{eq:incremental_metrics}
    \text{i\{METRIC\}} = \sum_{k \in K}\left[ \frac{k}{\sum_{k\in K} k} \text{\{METRIC\}}\left(\mathcal{X}^k, \mathcal{X}^k_{pgt}\right)\right]
\end{equation}
where $K$ are a set of keyframes which are taken every $m$ iterations, $\mathcal{X}^k$ is the solution computed at keyframe $k$, $\mathcal{X}_{pgt}^k$ is a pseudo-ground-truth defined as the iSAM2 solution using only inliers at keyframe $k$, and METRIC is a stand in for one of ATE, Precision, or Recall. Precision and Recall are computed from measurement classifications where, for example, "True Positive" corresponds to an inlier that is correctly classified. For methods that do not explicitly classify measurements, we define an inlier as having a $\chi^2$ error less than $0.95$. This value was also used in dataset generation to ensure synthetic measurements are actual outliers.

\subsection{Experiments}
\vspace{-2pt}
\subsubsection{Experiment 1 - Outlier Quantity} \label{exp:outlier_quantity}
\begin{figure*}[t!]
    \vspace{0.2cm}
    \centering
    \begin{subfigure}{0.49\linewidth}
      \centering
      \includegraphics[width=\linewidth, trim=0in 0.08in 0in 0in, clip]{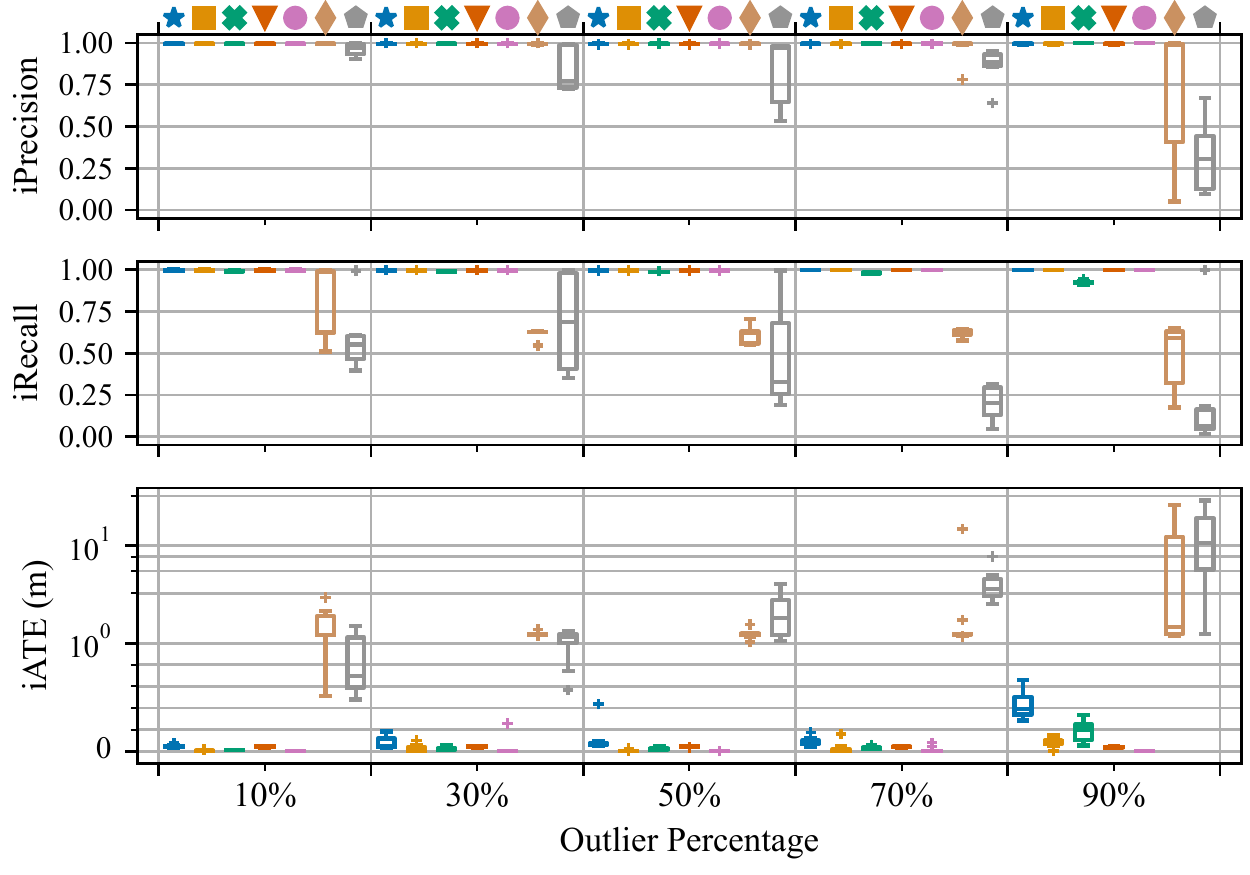}
      \caption{Dataset: Intel}
      \label{fig:exp1:intel}
    \end{subfigure}
    \begin{subfigure}{0.49\linewidth}
      \centering
      \includegraphics[width=\linewidth, trim=0in 0.08in 0in 0in, clip]{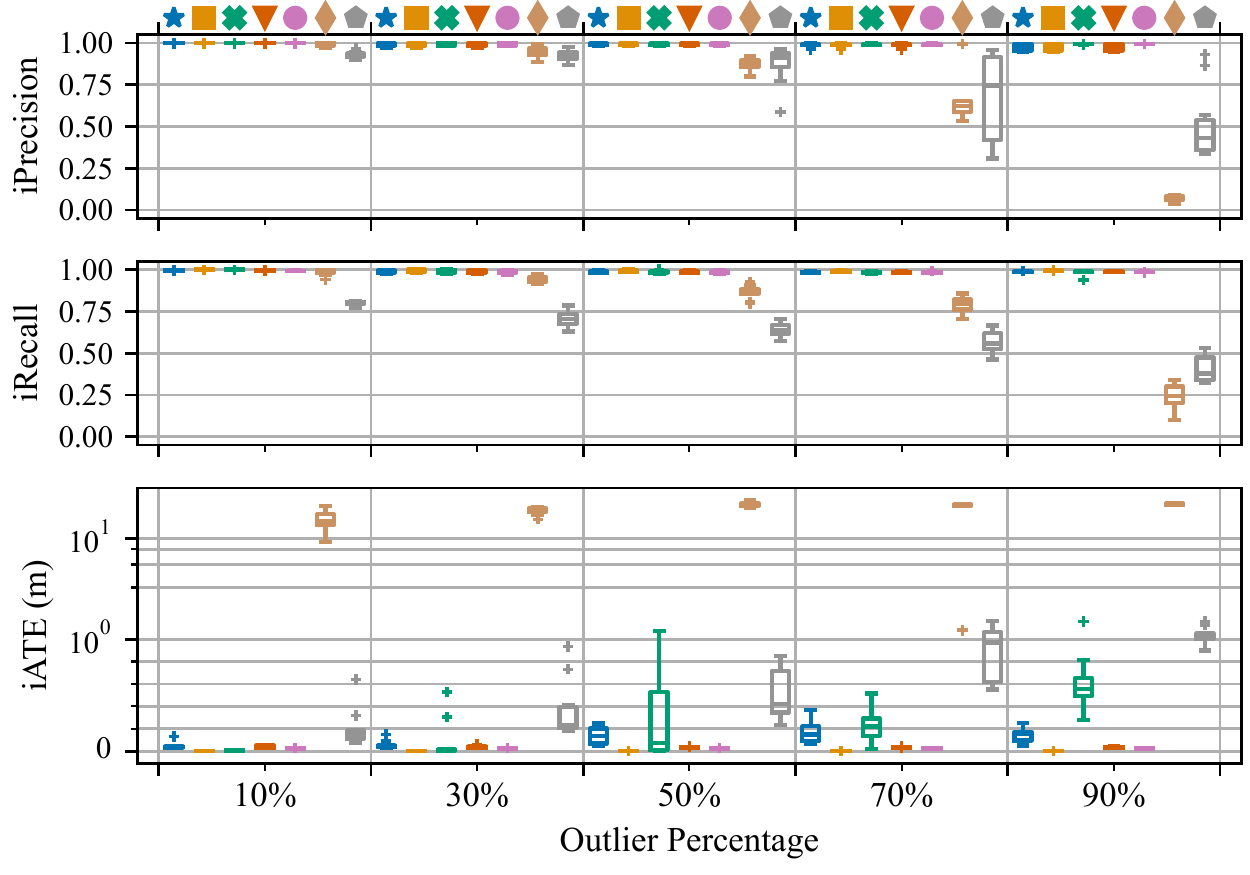}
      \caption{Dataset: CSAIL}
      \label{fig:exp1:csail}
    \end{subfigure}%
    \caption{
       Performance of our incremental algorithm riSAM (\SymRisam), prior batch methods GNC (\SymGnc) and PCM (\SymPcm), and prior incremental methods Geman-McClure (\SymGm), Max-Mixture (\SymMaxMix), Huber (\SymHuber), and Switchable Constraints (\SymSC) evaluated on the Intel (\ref{fig:exp1:intel}) and CSAIL (\ref{fig:exp1:csail}) datasets corrupted with outlier measurements. Our incremental method achieves comparable performance to its batch variant and performs well even with very large numbers of outlier measurements. \textit{Note: iATE plot uses linear scale up to $10^0$ and log scale above.}
    }
    \label{fig:exp_1_outlier_quantity}
    \vspace{-0.5cm}
\end{figure*}

We evaluate the robustness of the riSAM algorithm to different quantities of outliers. To provide continuity with prior works we evaluate on the MIT CSAIL Dataset~\cite{carlone_fast_2014} and the Intel Research Lab Dataset~\cite{carlone_angular_2014}. We generated outliers as a percentile of total loop-closures. To generate outliers we sample pairs of poses and add an identity loop-closure measurement between them. For each outlier percentile we generate ten random trials and report the incremental metrics via a box-and-whisker plot. The results from this experiment can be seen in Fig.~\ref{fig:exp_1_outlier_quantity}.

\subsubsection{Experiment 2 - Initialization} \label{exp:initialization}
\begin{figure}[h!]
    \centering
    \begin{subfigure}{0.4\linewidth}
      \centering
      \includegraphics[width=\linewidth]{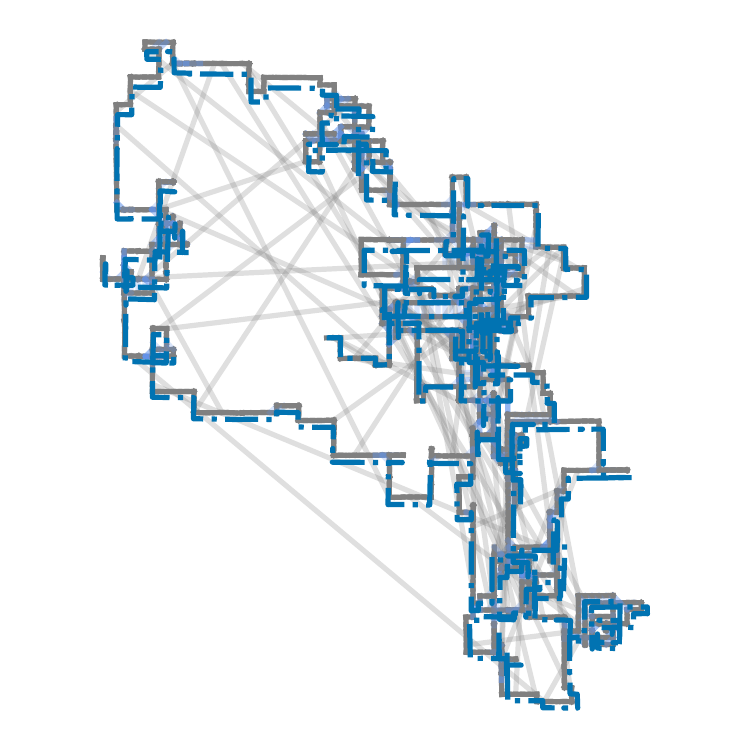}
      \vspace{-0.75cm}\caption{Small Noise: $\sigma_{\theta}=0.05^\circ$}
      \label{fig:example_gridworld_trajectories:small_noise}
    \end{subfigure}%
    \begin{subfigure}{0.4\linewidth}
      \centering
      \includegraphics[width=\linewidth]{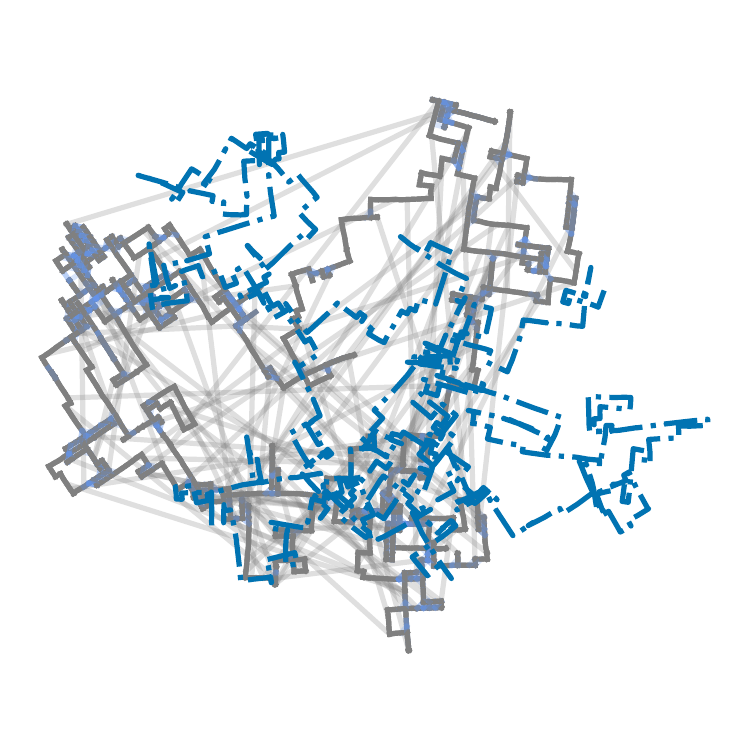}
      \vspace{-0.75cm}\caption{Large Noise: $\sigma_{\theta}=2.0^\circ$}
      \label{fig:example_gridworld_trajectories:large_noise}
    \end{subfigure}%
    \caption{Example random grid world trajectories. The trajectories' pseudo-ground-truths are shown in solid gray, while the odometric initializations are shown in dashed blue. Small odometric noise results in good initialization (\ref{fig:example_gridworld_trajectories:small_noise}) while larger odometric noise results in poor initialization (\ref{fig:example_gridworld_trajectories:large_noise}).}
    \label{fig:example_gridworld_trajectories}
\end{figure}

\begin{figure}[ht]
    \centering
    \includegraphics[width=\linewidth, trim=0in 0.05in 0in 0in, clip]{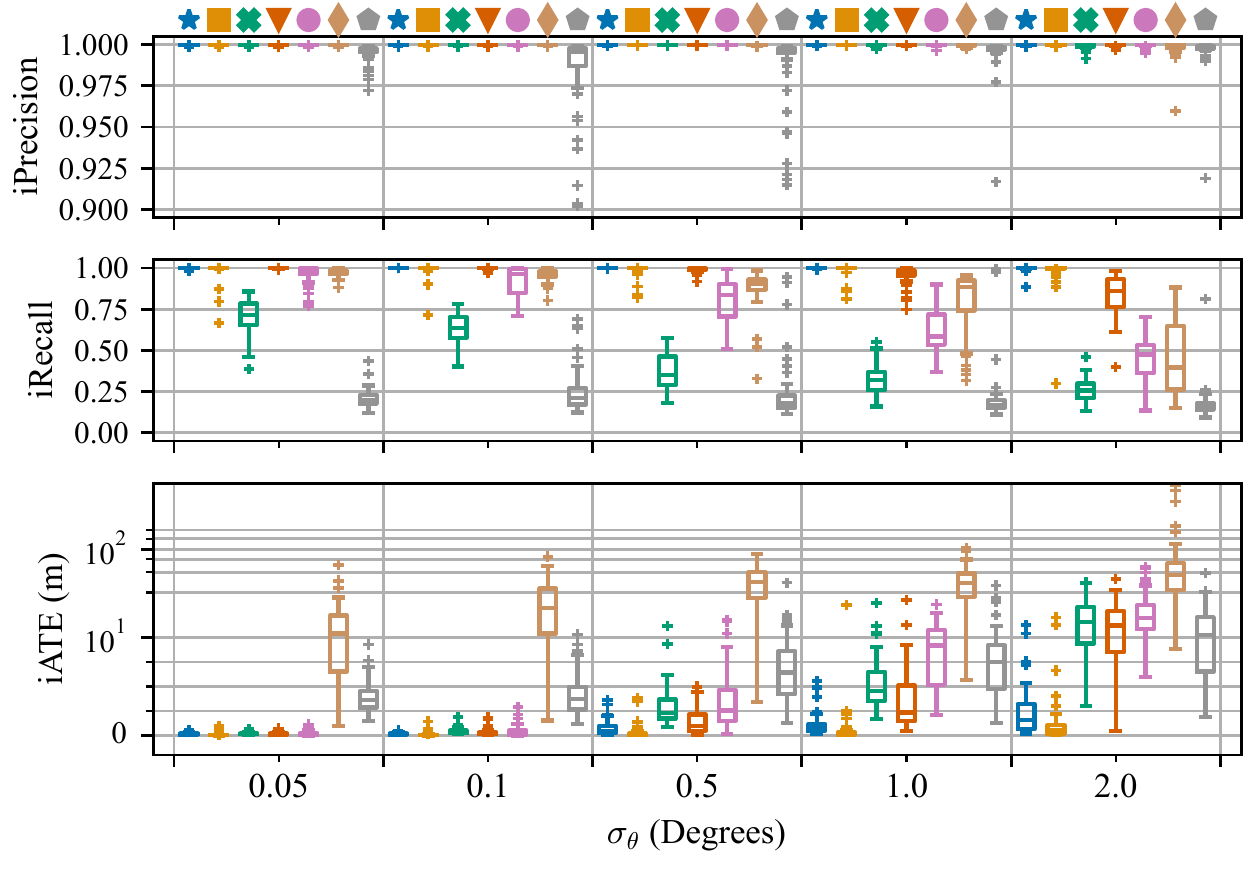}
    \caption{
        Performance of our incremental algorithm riSAM (\SymRisam), prior batch methods GNC (\SymGnc) and PCM (\SymPcm), and prior incremental methods Geman-McClure (\SymGm), Max-Mixture (\SymMaxMix), Huber (\SymHuber), and Switchable Constraints (\SymSC) evaluated on random grid world trajectories generated with increasing levels of odometric noise. Our method, riSAM, outperforms all others methods as the initialization quality decreases with larger odometric noise. \textit{Note: iATE plot has linear scale up to $10^2$ and log scale above.}
    }
    \label{fig:exp_2_initialization_results}
\end{figure}

We evaluate the robustness of riSAM to poor initialization. In the incremental SLAM problem we use the existing solution to initialize new variables. Under small amounts of noise this means new variables are initialized close to their ground truth value. However, under significant noise, variables are likely to be initialized far from their ground truth value. Therefore, by varying the amount of odometric noise we can evaluate how the different methods handle good (low noise) and poor (large noise) initialization. We generate random grid world trajectories like the examples shown in Fig.~\ref{fig:example_gridworld_trajectories} using different levels of odometric orientation noise. 
We generate 50 random trajectories for each noise level. Outliers are added during trajectory generation with probability $0.1$ at each step in which there is no inlier measurement. Results from this experiment can be found in Fig.~\ref{fig:exp_2_initialization_results}.

\subsubsection{Experiment 3 - Runtime} \label{exp:runtime}
\begin{figure}[t]
    \centering
    \includegraphics[width=\linewidth, trim=0in 0.05in 0in 0in, clip]{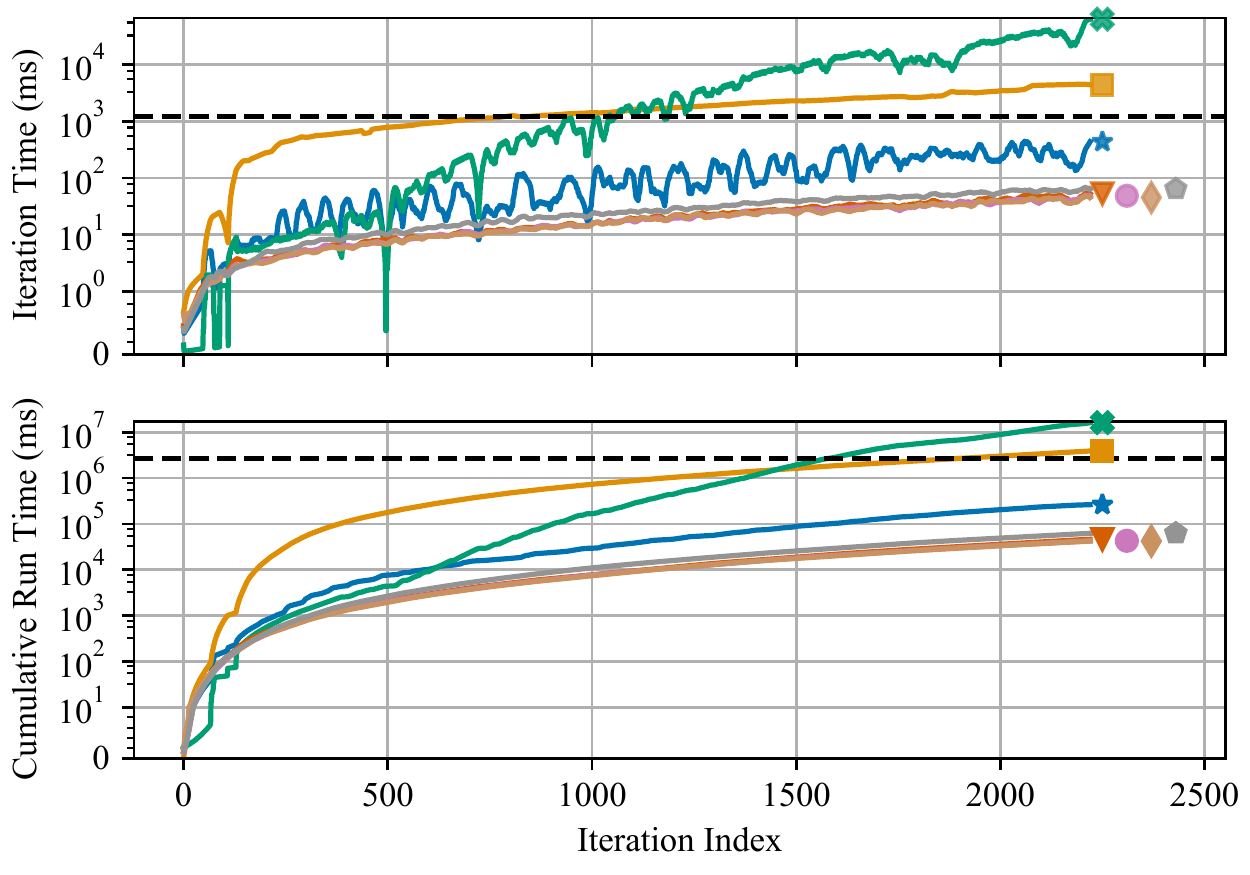}
    \caption{Runtime comparison between our incremental algorithm riSAM (\SymRisam), prior batch methods GNC (\SymGnc) and PCM (\SymPcm), and prior incremental methods Geman-McClure (\SymGm), Max-Mixture (\SymMaxMix), Huber (\SymHuber), and Switchable Constraints (\SymSC) on the Intel dataset. The batch methods do not meet real-time requirements (shown by horizontal dashed lines) while our algorithm and prior incremental methods satisfy online constraints. \textit{Note: Plots use a linear scale on the y-axis up to $10^0$ and $10^1$ respectively and log scale above those thresholds.}
    }
    \label{fig:exp_3_timing_results}
\end{figure}

In this experiment we evaluate the runtime performance of the riSAM algorithm. We do so using the Intel dataset as it has documented runtime information. We generate a single trial with $30\%$ outliers. We run each method five times on the same dataset and average results to reduce noise from CPU interrupts. We evaluate using per-iteration and total runtime. This experiment was run on a machine with an Intel Core i7-8665U processor. The results can be seen in Fig.~\ref{fig:exp_3_timing_results}.

\subsubsection{Experiment 4 - Dimensionality} \label{exp:dimensionality}
All previous experiments have used 2D SLAM problems. In this section we validate that the riSAM algorithm remains effective for 3D problems that real-world robots face. We report the performance of all methods from a single trial on the Sphere 2500 dataset~\cite{carlone_initialization_2015} with $10\%$ outliers. As we can see qualitatively in Fig.~\ref{fig:example_risam_solutions} and quantitatively in Tab.~\ref{tab:exp_4_dim_results} riSAM is able to perform well regardless of the dimensionality of the underlying problem.
\begin{table}[h]
    \centering
    \setlength\tabcolsep{4pt}
    \renewcommand{\arraystretch}{1.1}
    \begin{tabular}{|c|c|c|c|c|c|c|c|}
        \hline
                   & riSAM & GNC  & PCM$^*$ & GM   & Max-Mix & Huber & SC \\
        \hline
        iPrecision & 1.0   & 1.0  &  0.96   & 1.0  & 0.99    & 0.99  & 1.0 \\
        iRecall    & 0.99  & 0.97 &  0.99   & 0.98 & 0.02    & 0.23  & 0.32 \\
        iATE (m)   & 2.12  & 2.89 &  33.5   & 2.46 & 23.5    & 26.7  & 9.22 \\
        \hline
    \end{tabular}
    \vspace{-0.15cm}
    \caption{Evaluation of methods on the 3D Sphere 2500 dataset. riSAM generalizes across the dimensionality of the problem. $^*$PCM results computed on the first $63\%$ of keyframes as the method did not terminate in reasonable time.}
    \label{tab:exp_4_dim_results}
\end{table}
\subsection{Ablation Studies}
\vspace{-2pt}
In Sec.~\ref{sec:methodology:dogleg_line_search} we claim that that Dog-Leg Line Search is more effective than RISE~\cite{rosen_rise_2014} in riSAM and in Sec.~\ref{sec:methodology:risam} we claim that updating the per-factor $\mu_{init}$ improves long-term efficiency. In the following studies we validate these claims.

\subsubsection{Ablation 1 - Update Algorithm} In this study we evaluate our riSAM algorithm against a variant that uses the RISE algorithm~\cite{rosen_rise_2014}. We compare using a single trial of the Manhattan 3500 dataset with $30\%$ outliers~\cite{olson_fast_2006, carlone_angular_2014}. We report the incremental metrics in Table~\ref{tab:exp_5_ablation_1_update_alg_results} which show that the use of our novel Dog-Leg Line Search algorithm greatly improves riSAM's recall and trajectory error.

\begin{table}[H]
    \centering
    \begin{tabular}{|c|c|c|c|}
        \hline
        &  iPrecision & iRecall & iATE (m)\\
        \hline
        Ours (Alg.\ref{alg:dogleg_line_search}) & 1.0 & 0.99 & 0.56 \\
        RISE~\cite{rosen_rise_2014} & 0.99 & 0.87 & 11.09 \\
        \hline
    \end{tabular}
    \vspace{-0.15cm}
    \caption{Comparison between riSAM using RISE~\cite{rosen_rise_2014} and Dog-Leg Line Search (Alg.~\ref{alg:dogleg_line_search}). Dog-Leg Line Search produces better results by all metrics.}
    \label{tab:exp_5_ablation_1_update_alg_results}
\end{table}

\subsubsection{Ablation 2 - Initial Control Parameter Adjustment} In this study we compare our riSAM algorithm against a variant that omits updates to $\mu^i_{init}$. We evaluate both versions on the City10k dataset~\cite{isam}. Results are reported in Table~\ref{tab:exp_5_ablation_2_mu_inc_results} which shows that adjusting the per-factor $\mu^i_{init}$ does not degrade the performance of riSAM by any metric. Additionally, adjusting $\mu^i_{init}$ results in a \textbf{$\mathbf{24\%}$ faster} total runtime

\begin{table}[H]
    \centering
    \begin{tabular}{|c|c|c|c|}
        \hline
        &  iPrecision & iRecall & iATE (m) \\
        \hline
        Adjust $\mu^i_{init}$ & 1.0 & 0.99 & 0.22 \\
        Constant $\mu_{init}$ & 1.0 & 0.99 & 0.47 \\
        \hline
    \end{tabular}
    \vspace{-0.15cm}
    \caption{Comparison of riSAM with and without $\mu^i_{init}$ adjustment. Adjustment does not decrease performance of riSAM and provides a computational speed-up.}
    \label{tab:exp_5_ablation_2_mu_inc_results}
\end{table}

\section{Discussion and Future Work}\label{sec:discussion}
\vspace{-2pt}

 The results in Sec.~\ref{sec:experiments} demonstrate the capabilities of riSAM. In Fig.~\ref{fig:exp_1_outlier_quantity} and Fig.~\ref{fig:exp_2_initialization_results} we can see that riSAM is able to achieve iPrecision and iRecall statistics as good as batch GNC. riSAM often has slightly poorer iATE than GNC as intermediate results are sometimes taken when riSAM has not yet converged due to the approximation in Alg.~\ref{alg:efficient_gnc}. Further, in Fig.~\ref{fig:exp_3_timing_results} we observe that, unlike batch GNC, riSAM can achieve online efficiency on large scale problems. Looking holistically at robustness to both outliers and initialization, riSAM outperforms all other methods evaluated.

Some incremental methods (Switchable Constraints and Huber) perform poorly under all conditions as they can admit arbitrary influence from outliers. While this behavior was expected from the Huber kernel it is a surprise from Switchable Constraints which we expected to match the performance of a Geman-McClure kernel. The cause of this behavior is that, unlike M-Estimators, Switchable Constraints initializes switch variables to indicate that new measurements are inliers. This initialization effectively convexifies new measurements. However, unlike our method, the convexification is only done for new factors giving them disproportionate influence and resulting in poor performance.

Other incremental methods (Max-Mixtures and Geman-McClure) and surprisingly the batch method, PCM, appear to perform very well in the first experiment (Fig.~\ref{fig:exp_1_outlier_quantity}), when initialization is good, but break down when odometric drift results in poor initialization in the second experiment (Fig.~\ref{fig:exp_2_initialization_results}). For the kernel methods this behavior is a result of the fact that they are effectively asymptotically constant kernels for the given parameters and thus sensitive to initialization. For PCM, we hypothesize that this behavior is caused by the assumptions made by the algorithm for tractability.

Currently, riSAM's largest drawback is its dependence on user defined parameters. Future work should investigate the affect of these parameters on the algorithm's behavior. Future work could also include extending riSAM for distributed applications and running field-trials for additional validation.


\newpage
\bibliographystyle{IEEEtran} 
\bibliography{refs}

\end{document}